\documentclass[letterpaper,10pt,conference]{ieeeconf}

\IEEEoverridecommandlockouts
\usepackage[T1]{fontenc}
\usepackage{amsmath}
\usepackage{amssymb}
\usepackage{algorithm}
\usepackage{algpseudocode}
\usepackage{booktabs}
\usepackage{tabularx}
\usepackage[nocompress]{cite}
\usepackage{graphicx}
\usepackage{tikz}
\usetikzlibrary{arrows.meta,calc,positioning}
\usepackage{dblfloatfix}
\usepackage{placeins}

\makeatletter
\let\NAT@parse\undefined
\renewcommand{\@IEEEfigurecaptionsepspace}{\vskip 3pt\nointerlineskip}
\makeatother
\usepackage[hidelinks]{hyperref}

\title{\LARGE \bf AlphaNIST: Emergent Assembly Skills from Binary Task Rewards}

\author{%
    \authorblockN{Octi Zhang$^{1}$, Iretiayo Akinola$^{1}$, Rosario Scalise$^{2}$,
    Mateo Guaman Castro$^{2}$, Patrick Yin$^{1,2}$\\
    Tobias Widmer$^{1}$, Miles Macklin$^{1}$, Chris Amevor$^{1}$,
    Dylan Turpin$^{1}$, Dirk Van Gelder$^{1}$\\
    Jan Carius$^{1}$, Ankur Handa$^{1}$, Gavriel State$^{1}$,
    Yunrong Guo$^{1}$, and Yashraj Narang$^{1}$}
    \authorblockA{$^{1}$NVIDIA \qquad $^{2}$University of Washington}
}

\makeatletter
\IEEEaftertitletext{%
    \noindent\begin{minipage}{\textwidth}
        \centering
        \includegraphics[width=\linewidth]{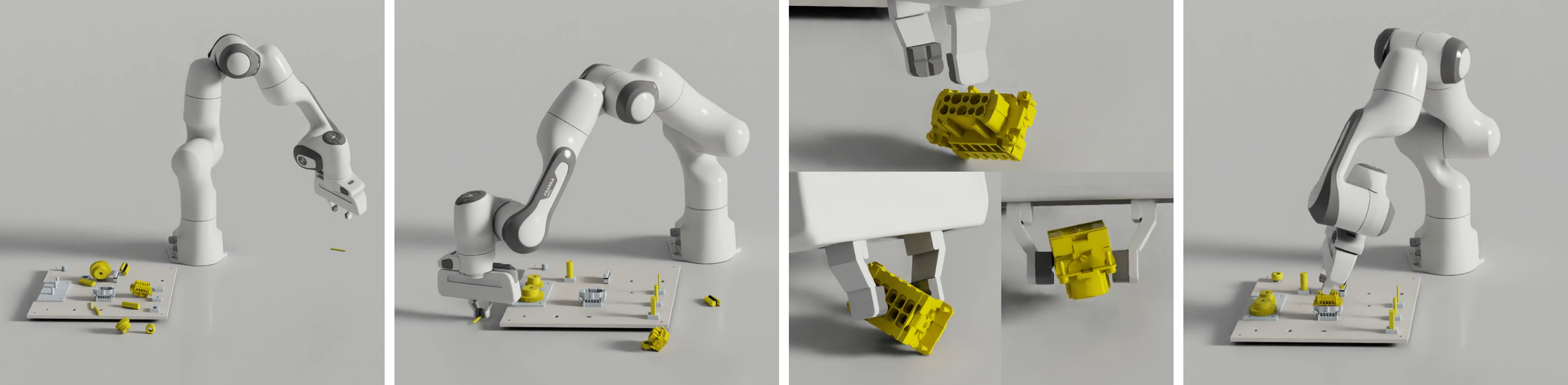}
        \endgraf\nointerlineskip
        \setlength{\abovecaptionskip}{3pt}
        \def\@captype{figure}
        \caption{
        \textbf{Emergent multi-component assembly with a single policy}. Trained with a binary task reward, AlphaNIST learns to select, grasp, reorient, and assemble diverse components, while autonomously sequencing multiple assemblies within a single episode. Shown are snapshots from a ten-component assembly sequence.}
        \label{fig:assembly-hero}
    \end{minipage}
    \vspace{0.5\baselineskip}%
}
\makeatother

\begin{document}

\maketitle
\thispagestyle{empty}
\pagestyle{empty}

\begin{abstract}
The NIST Assembly Task Board \#1 remains a challenging benchmark for learning-based methods in robotic manipulation.
Over short horizons, it demands precise, contact-rich control of small, tight-fitting parts.
Over long horizons, it requires recovery and coordination across extended action sequences.
Demonstrations, task decomposition, and dense progress rewards have enabled research advances by providing guidance on how to assemble.
But how much of the strategy must be specified?
The progression from AlphaGo to AlphaGo Zero showed that RL at scale can discover superhuman strategies without prescribing how to win.
AlphaNIST asks whether a sparse binary task reward and task diversity admitted by this task-agnostic reward can similarly let a robot discover how to assemble.
In simulation, we train a \textit{single} policy from scratch across all 20 assembly tasks on the NIST Assembly Task Board \#1, reaching \textbf{98.2\%} overall training success without demonstrations, progress shaping, scripted primitives, or per-task tuning.
The framework also trains policies to complete ten assembly pairs in a single episode without a prescribed order, with task success rewarded \textit{only when} all ten pairs are simultaneously complete.
The learned policies discover dexterous and planning-like behaviors: choosing grasps for later assembly, using surrounding geometry to reorient parts, and preparing one part, assembling another, then returning to finish the first without an explicit planner.
For videos, see our website: \url{https://alpha-nist.github.io/}

\end{abstract}

\section{Introduction}
\label{sec:introduction}
Robotic assembly is a longstanding challenge, and the NIST Assembly Task Boards are a persistent benchmark~\cite{kimble2020benchmarking}; to the best of our knowledge, no learning-based approaches have solved all tasks on any of the boards. Assembly Task Board \#1 (ATB1) consists of 20 \textit{assembly pair} tasks (a part and its fixed mating geometry on the board) that are relevant for manufacturing, including peg insertion, gear meshing, electrical connector insertion, and nut threading.
Solving this task board is exceptionally challenging for three key reasons:
\textbf{(1)} The sequence for solving the tasks might be unspecified, and not all sequences may be efficient or valid (e.g., solving task N may obstruct the robot from accessing the parts in task N + 1).
\textbf{(2)} Each task is long-horizon, encompassing grasping, transport, and
contact-rich assembly~\cite{narang2022factory}.
Parts may begin in a wide range of poses and require multiple
reorientations before grasping.
During transport, parts may need further reorientation in-hand or
through extrinsic contact.
During assembly, parts may slip, become misaligned, or
jam during insertion or threading~\cite{narang2022factory}, requiring
robust recovery and reexecution.
\textbf{(3)} Solving the tasks requires precise control during complex contact interactions. The parts themselves are small (as small as 4\,mm in diameter), have high geometric detail (e.g., helical threads), and have tight clearances between mating surfaces (as low as zero).
Tasks such as nut threading involve learning helical motion during frictional contact between multiple mating surfaces, during which a nut can tilt, cross-thread, or jam instead of advancing along the bolt~\cite{narang2022factory}.

Learning-free approaches typically rely on planning and strong human priors:
assembly sequence planning~\cite{lozano1993assembly}; customized fixtures,
adapters, and tooling~\cite{kimble2020benchmarking,lian2021benchmarking}; engineered
motion and force primitives~\cite{morgan2023contact}; and tuned admittance,
impedance, or force-motion controllers~\cite{lian2021benchmarking,morgan2023contact}.

Learning-based approaches, on the other hand, can relax some of these constraints, but have shown progress only within a limited scope.
Most efforts have assumed a fixed assembly sequence, heavily constrained the initial poses of parts (e.g., beginning in an already-grasped state)~\cite{luo2021robust}, assumed access to accurate pose estimates, learned specialist rather than multi-task policies~\cite{narang2022factory}, addressed only a subset of tasks on the board~\cite{luo2021robust}, and/or achieved only low-to-moderate success rates by industrial standards.
Success-Guided Sampling (SGS)~\cite{zhang2026balanceddiet}, for example, solves tasks one at a time, but requires a separate policy per task and is considerably less compute-efficient.

We draw inspiration from the landmark AlphaGo series in reinforcement
learning, which surpassed the strongest human Go players.
The progression from AlphaGo to AlphaGo Zero 
showed that learning from a terminal task objective
could encourage superhuman decision-making over long horizons without human demonstrations~\cite{silver2016alphago,silver2017alphagozero}.
In this paper,  we ask whether a similar approach is possible for robotics: can sparse binary task rewards, combined with appropriate training strategies, enable a robot to solve sequences of long-horizon, contact-rich manipulation tasks, perhaps leveraging dexterous behaviors that no human could feasibly specify or demonstrate?
To make learning from sparse task rewards more tractable, we take a seemingly
counterintuitive step: train jointly across heterogeneous assembly pairs differing in geometries, tolerances, and assembly requirements (a nut and its bolt, a gear and its gearshaft, a peg and its hole) from a broad coverage of
initial states, \textbf{converting MDP diversity into a resource rather than a burden}.

\begin{figure}[t]
    \centering
    \includegraphics[width=\columnwidth]{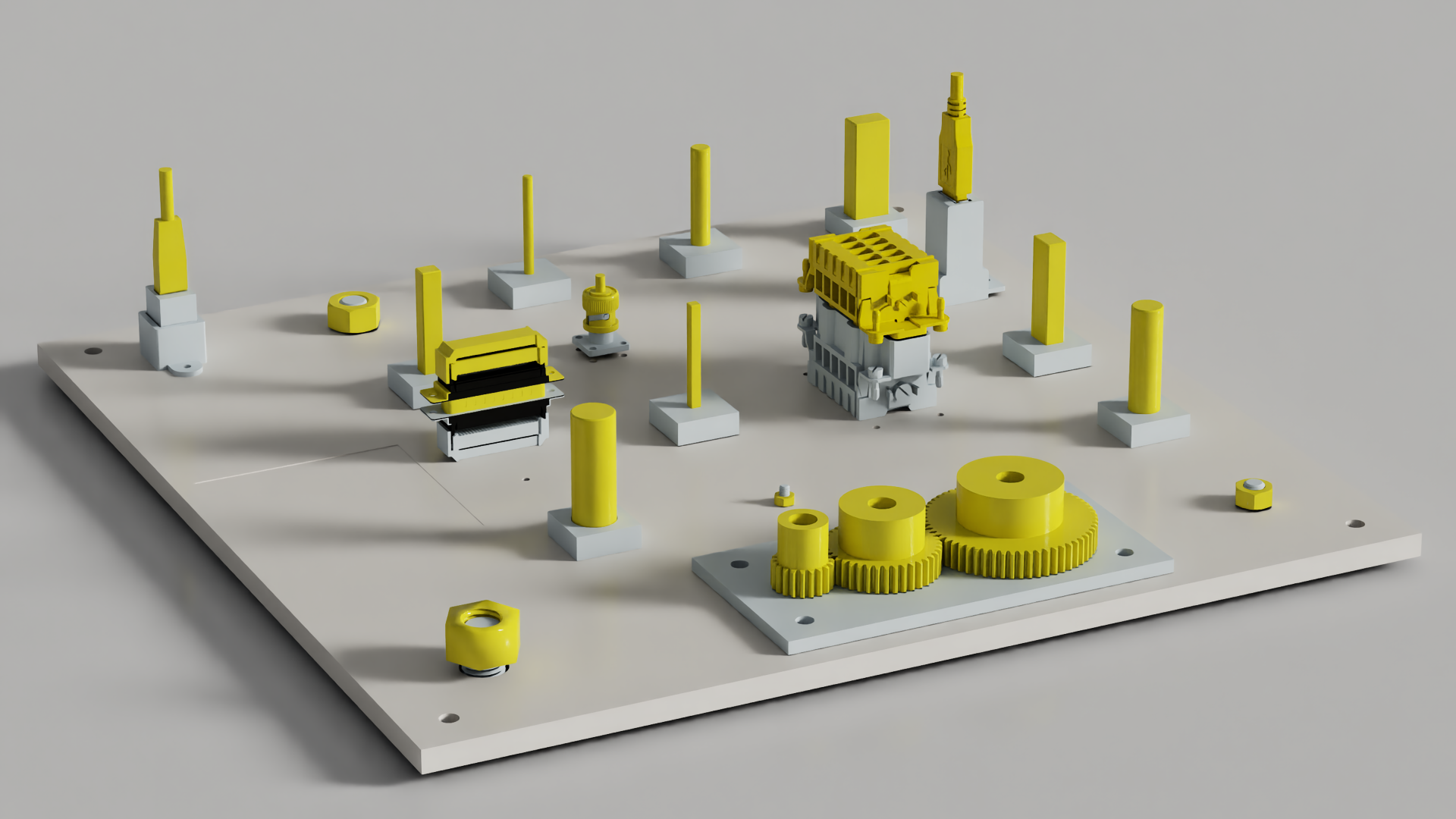}
    \caption{\textbf{The complete NIST Assembly Task Board \#1 in simulation.}
    Yellow indicates the 20 movable parts in their fully-assembled
    states.}
    \label{fig:nistboard-overview}
\end{figure}

Guided by this insight, we develop AlphaNIST, a learning framework that
trains a \textit{single policy} under sparse binary task rewards across all 20 tasks
on the NIST Assembly Task Board 1 (Fig. \ref{fig:nistboard-overview}).
The policy handles an exceptional range of initial part poses, learns
from point-cloud inputs without first training specialist policies,
and achieves a 98.2\% overall success rate.
Furthermore, the same framework can train policies to complete 10 assembly pairs in
sequence, with a task reward issued only when all pairs are assembled with
\textit{no prescribed order}.
In other words, the policy exhibits planning-like behavior.

Our contributions are the following:
\begin{enumerate}
    \item \textbf{A highly-effective multi-task RL framework} consisting of a novel combination of the following components: A) generation of a cache of assembly-relevant reset states~\cite{yin2026omnireset} over heterogeneous assembly pairs, B) success-guided sampling within the cache~\cite{zhang2026balanceddiet}, C) a unified sparse binary task reward, D) multi-GPU training at scale. These components are supported by careful ablations. All tasks share one training configuration, without per-task tuning or demonstrations.
    \item \textbf{The first multi-task RL policy capable of solving all 20 tasks on the NIST Assembly Task Board 1}. The policy takes point clouds as input; initial part poses are sampled from a wide range, including adverse conditions (e.g., upside-down); and each task can be solved with near-perfect success rates. Positive transfer is observed, where multi-task policies can solve the 20 tasks with higher success rate than individual specialists, including three tasks that the specialists struggle to solve.
    \item \textbf{The first assembly policy capable of solving up to 10 diverse assembly pairs in sequence}, without an explicit hierarchy or prescribed task order. Thus, this work demonstrates that sequencing behavior can emerge within a single low-level assembly policy.
\end{enumerate}

We observe a repertoire of unique emergent behaviors including dexterous skills such as pushing and sliding (to singulate parts), in-hand pivoting, multiple sequential regrasps, extrinsic contact (to reorient parts prior to insertion), and single-gripper nut threading.
The single policy also exhibits long-horizon planning-like behavior such as bringing parts into reach, moving obstructing parts aside, and leaving a task partially-finished and then returning to it later. None of these skills and behaviors were specified.
We encourage readers to watch our supplementary videos \footnote{https://alpha-nist.github.io/} for additional examples of these emergent behaviors.
We plan to release our code following the review and revision period.

\section{Related Work}
\label{sec:related-work}
\subsection{Scalable Learning for Contact-Rich Assembly}
Parallel physics simulation supports experience collection at scale for
on-policy methods such as PPO~\cite{mittal2025isaaclab, zakka2026mjlab},
with applications ranging from dexterous in-hand manipulation
\cite{andrychowicz2020dexterous, qi2023general, singh2024dextrah, handa2023dextreme}
to industrial assembly.
GPU-accelerated physics engines such as PhysX~\cite{nvidiaphysx} and
Newton~\cite{newton2025} support large-scale parallel simulation for robot learning.
For industrial assembly,
Factory~\cite{narang2022factory} demonstrated real-time or faster simulation of
1,000 simultaneous nut-and-bolt interactions in Isaac Gym, later incorporated into Isaac Lab. 
AlphaNIST uses this
simulation throughput to collect experience across heterogeneous assembly pairs.

OmniReset~\cite{yin2026omnireset} addressed state coverage through diverse,
manipulation-centric resets, demonstrating long-horizon on-policy RL with fixed
algorithm hyperparameters and no curricula or human demonstrations.
Success-Guided Sampling (SGS)~\cite{zhang2026balanceddiet} addressed experience
allocation by concentrating simulation near the policy's changing capability
frontier, scaling beyond one million parallel environments using sparse rewards.
In its reported assembly experiments, SGS trains a separate policy for
each single task. AlphaNIST retains SGS's sampling rule but trains one
multi-task policy across all 20 ATB1 tasks under a common binary task reward.
Our experiments show that this shared policy learns assemblies that
remain unsolved in single-task training
(Fig.~\ref{fig:heterogeneity-ablation}).

\subsection{Multi-Task Reinforcement Learning}
Multi-task RL aims to improve learning by sharing experience across tasks.
However, standard joint-training baselines have achieved lower success
rates than separately trained policies on multi-task manipulation
benchmarks~\cite{yu2020pcgrad}.
One approach to obtaining a single multi-task policy is to distill
multiple separately trained specialists into a shared
policy~\cite{tang2024automate}.
Other approaches address the difficulty of joint training directly.
PopArt~\cite{hessel2019popart} addresses differences in reward scales
with a separate value output and normalization statistics for each task,
while PCGrad~\cite{yu2020pcgrad} removes conflicting components of task
gradients before combining them.
However, Kurin et al.~\cite{kurin2022unitary} show that shared-network
baselines with standard regularization and stabilization can match
specialized multi-task optimizers.

AlphaNIST takes a simple approach to multi-task learning by exploiting
the shared structure of assembly: threading, gear meshing, and insertion
all receive the same binary task reward when assembly error falls below
a common threshold. Under this shared objective, we train one
geometry-conditioned policy and one value function with standard PPO
at scale, without task-specific value heads, gradient projection,
or specialist pretraining.

\subsection{Learning-Based Robot Assembly}
Prior learning-based assembly systems make exploration tractable by shortening
the task horizon or restricting initialization to staged, pre-grasped, or
local-insertion problems
\cite{schoettler2020industrial,florensa2017reverse}.
Factory~\cite{narang2022factory} trains separate pick, place, and screw
subpolicies, while AutoMate~\cite{tang2024automate} combines
assembly-by-disassembly, imitation learning, and curriculum-based RL to obtain
specialist and generalist assembly policies.
Refinery~\cite{tang2025refinery} improves pretrained assembly specialists
through Bayesian-optimization-guided sampling of initial conditions during
fine-tuning and Gaussian-mixture-model-based initialization at deployment.
The refined policies can then be chained to complete multi-part assemblies.
For longer sequences,
ARCH~\cite{sun2025arch} uses a high-level imitation-learned policy to select from
a library of RL and model-based primitives. These methods structure learning
through initialization, demonstrations, or prescribed subskills.
OmniReset~\cite{yin2026omnireset} uses a generic dense objective with fixed
weights, including gripper--object and object--goal distance terms. Zhang et al.~\cite{zhang2026balanceddiet} present the first end-to-end RL solution to individual tasks from the NIST ATB1, but retain task-specific
auxiliary grasp- and gripper-related shaping. Such
terms can aid exploration and discourage reward exploitation, but their relative
influence can change with object size, geometry, and assembly motion. Even with
fixed coefficients, a new task family requires checking that the shaped reward
remains aligned with successful assembly.

AlphaGo Zero provides a broader motivation: terminal win/loss and
search-generated policy targets support strategy discovery without human game
data or intermediate reward shaping~\cite{silver2017alphagozero}. AlphaNIST asks
whether binary task success can similarly support strategy discovery in continuous,
contact-rich control, with one shared policy learning heterogeneous
assemblies end to end.

\section{Problem Formulation}
\label{sec:problem-formulation}
\subsection{Task Definition and MDP}
\label{sec:task_mdp}
We formulate all 20 ATB1 assembly tasks~\cite{kimble2020benchmarking}
as a multi-task reinforcement-learning problem with one shared policy.

Let $e \in \mathcal{E}$ denote an assembly task, with $|\mathcal{E}|=20$.
Each task corresponds to one \textit{assembly pair}: a movable \textit{part}, its \textit{mating geometry}, and a fully-assembled configuration.

We model the problem as a finite-horizon MDP,
\begin{equation}
    \mathcal{M}=(\mathcal{S},\mathcal{A},p,r,\gamma,T,\rho),
\end{equation}
where $p(s_{t+1}\mid s_t,a_t)$ is the transition kernel and $\rho$
is the joint distribution over task identities and initial states.
At the beginning of each episode, $(e,s_0)\sim\rho$.
A shared policy $\pi_\theta(a_t\mid o_t)$ acts on partial observations
$o_t$ of the simulator state $s_t$ and maximizes
\begin{equation}
    J(\theta)
    =
    \mathbb{E}_{\rho,p,\pi_\theta}
    \left[
        \sum_{t=0}^{T-1}\gamma^t r_t
    \right].
    \label{eq:objective}
\end{equation}
Task variation is represented through surface point clouds in $o_t$, 
rather than an explicit discrete task identifier.

\section{Method}
\label{sec:method}
\subsection{AlphaNIST Training Framework}

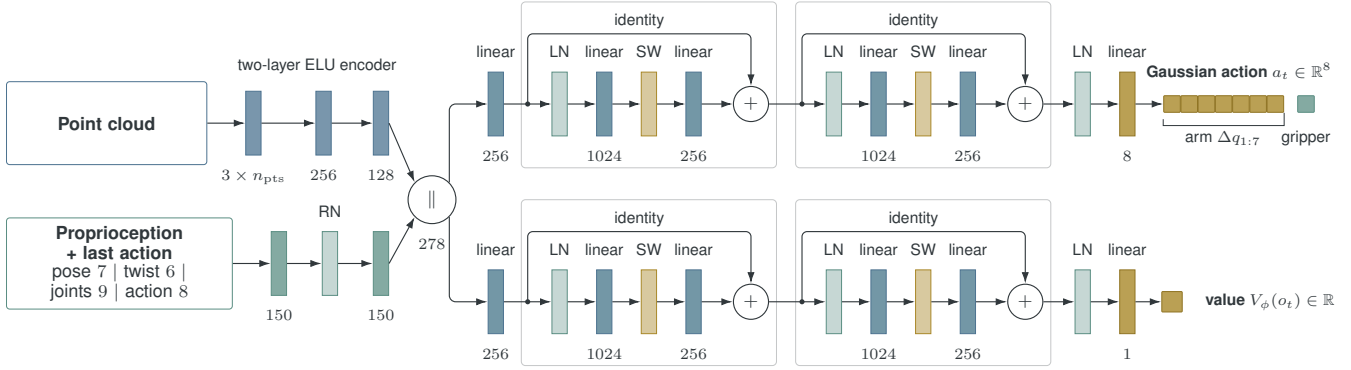
\begin{figure*}[t]
    \centering
    \resizebox{\textwidth}{!}{\begin{tikzpicture}[
    x=1mm,
    y=1mm,
    font=\footnotesize\sffamily,
    flow/.style={-{Latex[length=1.8mm,width=1.15mm]}, draw=graphite,
        line width=0.5pt},
    source/.style={rounded corners=0.5mm, line width=0.45pt, align=center,
        inner sep=1.2mm},
    layer/.style={line width=0.4pt, minimum width=2.4mm,
        minimum height=10mm, inner sep=0pt, align=center},
    annotation/.style={font=\scriptsize\sffamily, text=graphite, align=center}
]
    \definecolor{graphite}{RGB}{48,55,59}
    \definecolor{geomblue}{RGB}{77,115,144}
    \definecolor{stateblue}{RGB}{91,142,131}
    \definecolor{networkblue}{RGB}{115,151,166}
    \definecolor{networkgold}{RGB}{169,136,41}
    \tikzset{every node/.style={text=graphite}}

    \node[source, draw=geomblue, fill=white, text width=29mm,
        minimum height=13mm] (geometry) at (15,12)
        {\textbf{Point cloud}};
    \node[source, draw=stateblue, fill=white, text width=33mm,
        minimum height=12mm] (history) at (17,-10)
        {\textbf{Proprioception + last action}\\[-1pt]
        pose $7$ $\mid$ twist $6$ $\mid$ joints $9$ $\mid$ action $8$};

    \node[layer, draw=geomblue, fill=geomblue!75]
        (pc-in) at (38,12) {};
    \node[layer, draw=geomblue, fill=geomblue!75]
        (pc-hidden) at (49,12) {};
    \node[layer, draw=geomblue, fill=geomblue!75]
        (pc-out) at (58,12) {};
    \node[annotation] at (48,21) {two-layer ELU encoder};
    \node[annotation, below=1.0mm of pc-in] {$3\times n_{\mathrm{pts}}$};
    \node[annotation, below=1.0mm of pc-hidden] {$256$};
    \node[annotation, below=1.0mm of pc-out] {$128$};
    \draw[flow] (geometry.east) -- (pc-in.west);
    \draw[flow] (pc-in.east) -- (pc-hidden.west);
    \draw[flow] (pc-hidden.east) -- (pc-out.west);

    \node[layer, draw=stateblue, fill=stateblue!75]
        (state-in) at (42,-10) {};
    \node[layer, draw=stateblue, fill=stateblue!35]
        (state-norm) at (50,-10) {};
    \node[annotation, above=1.0mm of state-norm] {RN};
    \node[layer, draw=stateblue, fill=stateblue!75]
        (state-out) at (58,-10) {};
    \node[annotation, below=1.0mm of state-in] {$150$};
    \node[annotation, below=1.0mm of state-out] {$150$};
    \draw[flow] (history.east) -- (state-in.west);
    \draw[flow] (state-in.east) -- (state-norm.west);
    \draw[flow] (state-norm.east) -- (state-out.west);

    \node[circle, draw=graphite, fill=white, line width=0.5pt,
        minimum size=7.5mm, inner sep=0pt] (concat) at (66,0) {$\Vert$};
    \node[annotation, below=1.0mm of concat] {$278$};
    \draw[flow] (pc-out.east) -- (concat.north west);
    \draw[flow] (state-out.east) -- (concat.south west);

    \foreach \branch/\yy in {actor/15,critic/-16} {
        \node[layer, draw=geomblue, fill=networkblue]
            (projection-\branch) at (76,\yy) {};
        \node[annotation, above=1.0mm of projection-\branch] {linear};
        \node[annotation, below=1.0mm of projection-\branch] {$256$};

        \foreach \block/\offset in {1/0,2/43} {
            \path[draw=graphite!28, fill=white,
                line width=0.35pt, rounded corners=0.65mm]
                ({80+\offset},{\yy-10}) rectangle ({120+\offset},{\yy+16});
            \coordinate (res-in-\branch-\block) at ({81+\offset},\yy);
            \node[layer, draw=stateblue, fill=stateblue!35]
                (layer-norm-\branch-\block) at ({86+\offset},\yy) {};
            \node[annotation, above=1.0mm of layer-norm-\branch-\block] {LN};
            \node[layer, draw=geomblue, fill=networkblue]
                (expand-\branch-\block) at ({93+\offset},\yy) {};
            \node[annotation, above=1.0mm of expand-\branch-\block] {linear};
            \node[annotation, below=1.0mm of expand-\branch-\block] {$1024$};
            \node[layer, draw=networkgold, fill=networkgold!45]
                (swish-\branch-\block) at ({100+\offset},\yy) {};
            \node[annotation, above=1.0mm of swish-\branch-\block] {SW};
            \node[layer, draw=geomblue, fill=networkblue]
                (contract-\branch-\block) at ({107+\offset},\yy) {};
            \node[annotation, above=1.0mm of contract-\branch-\block] {linear};
            \node[annotation, below=1.0mm of contract-\branch-\block] {$256$};
            \node[circle, draw=graphite, fill=white,
                line width=0.5pt, minimum size=5.5mm, inner sep=0pt]
                (res-add-\branch-\block) at ({116+\offset},\yy) {$+$};
            \draw[flow] (layer-norm-\branch-\block.east) -- (expand-\branch-\block.west);
            \draw[flow] (expand-\branch-\block.east) -- (swish-\branch-\block.west);
            \draw[flow] (swish-\branch-\block.east) -- (contract-\branch-\block.west);
            \draw[flow] (contract-\branch-\block.east) -- (res-add-\branch-\block.west);
            \draw[flow, rounded corners=0.8mm]
                (res-in-\branch-\block) -- ++(0,11mm) --
                node[above=0.4mm, annotation, fill=white, inner sep=0.4mm] {identity}
                ({116+\offset},{\yy+11}) -- (res-add-\branch-\block.north);
            \fill[graphite] (res-in-\branch-\block) circle[radius=0.4mm];
        }
        \draw[flow] (projection-\branch.east) -- (layer-norm-\branch-1.west);
        \draw[flow] (res-add-\branch-1.east) -- (layer-norm-\branch-2.west);

        \node[layer, draw=stateblue, fill=stateblue!35]
            (out-norm-\branch) at (168,\yy) {};
        \node[annotation, above=1.0mm of out-norm-\branch] {LN};
        \node[layer, draw=networkgold!85!black, fill=networkgold!80]
            (head-\branch) at (175,\yy) {};
        \draw[flow] (res-add-\branch-2.east) -- (out-norm-\branch.west);
        \draw[flow] (out-norm-\branch.east) -- (head-\branch.west);
    }

    \draw[flow, rounded corners=1mm]
        (concat.north east) |- (projection-actor.west);
    \draw[flow, rounded corners=1mm]
        (concat.south east) |- (projection-critic.west);

    \node[annotation, above=1.0mm of head-actor] {linear};
    \node[annotation, below=1.0mm of head-actor] {$8$};
    \foreach \j [evaluate=\j as \x using 182+(\j-1)*2.7] in {1,...,7} {
        \node[draw=networkgold!85!black, fill=networkgold!80, line width=0.45pt,
            rounded corners=0.15mm, minimum size=2.6mm, inner sep=0pt]
            (arm-\j) at (\x,15) {};
    }
    \node[draw=stateblue, fill=stateblue!75, line width=0.45pt,
        rounded corners=0.15mm, minimum size=2.6mm, inner sep=0pt]
        (gripper) at (203,15) {};
    \node[annotation, font=\scriptsize\sffamily\bfseries] at (192.5,20.2)
        {Gaussian action $a_t\in\mathbb{R}^{8}$};
    \draw[flow] (head-actor.east) -- (arm-1.west);
    \draw[draw=graphite, line width=0.4pt]
        ([yshift=-1mm]arm-1.south west) -- ++(0,-1mm) -|
        ([yshift=-1mm]arm-7.south east) -- ++(0,1mm);
    \node[annotation, text=graphite] at (190,9.6)
        {arm $\Delta q_{1:7}$};
    \node[annotation, text=graphite] at (203,9.6)
        {gripper};

    \node[annotation, above=1.0mm of head-critic] {linear};
    \node[annotation, below=1.0mm of head-critic] {$1$};
    \node[draw=networkgold!85!black, fill=networkgold!80, line width=0.45pt,
        rounded corners=0.15mm, minimum size=3.2mm, inner sep=0pt]
        (value-cell) at (182,-16) {};
    \node[annotation, font=\scriptsize\sffamily\bfseries,
        text=graphite, anchor=west] at (186,-16)
        {value $V_\phi(o_t)\in\mathbb{R}$};
    \draw[flow] (head-critic.east) -- (value-cell.west);
\end{tikzpicture}}
    \caption{\textbf{Policy architecture.} 
    Actor and critic have the same architecture with separate
    parameters. Inputs are drawn once; each network contains two consecutive
    residual blocks. RN: running normalization; LN: layer normalization;
    SW: Swish activation; $n_{\mathrm{pts}}$: number of input points.}
    \label{fig:policy-architecture}
\end{figure*}

AlphaNIST trains one policy across all 20 assembly tasks with standard PPO
and a binary task reward for completed assembly.
Its training distribution combines diverse cached
reset states from OmniReset~\cite{yin2026omnireset} with beta-distributed success-guided sampling (SGS)~\cite{zhang2026balanceddiet}, while a common binary task
reward defines successful assembly across all tasks. The three components
play distinct roles:

\[
\begin{array}{r@{\quad}l}
\textbf{Reset bank:}    & \textit{Which states are available?}\\
\textbf{Beta sampling:} & \textit{Which states are sampled?}\\
\textbf{Binary task reward:} & \textit{What counts as success?}
\end{array}
\]

\subsection{Reset Coverage and Experience Allocation}
\label{sec:success_guided_sampling}

\begin{figure}[!tb]
    \centering
    \includegraphics[width=\columnwidth]{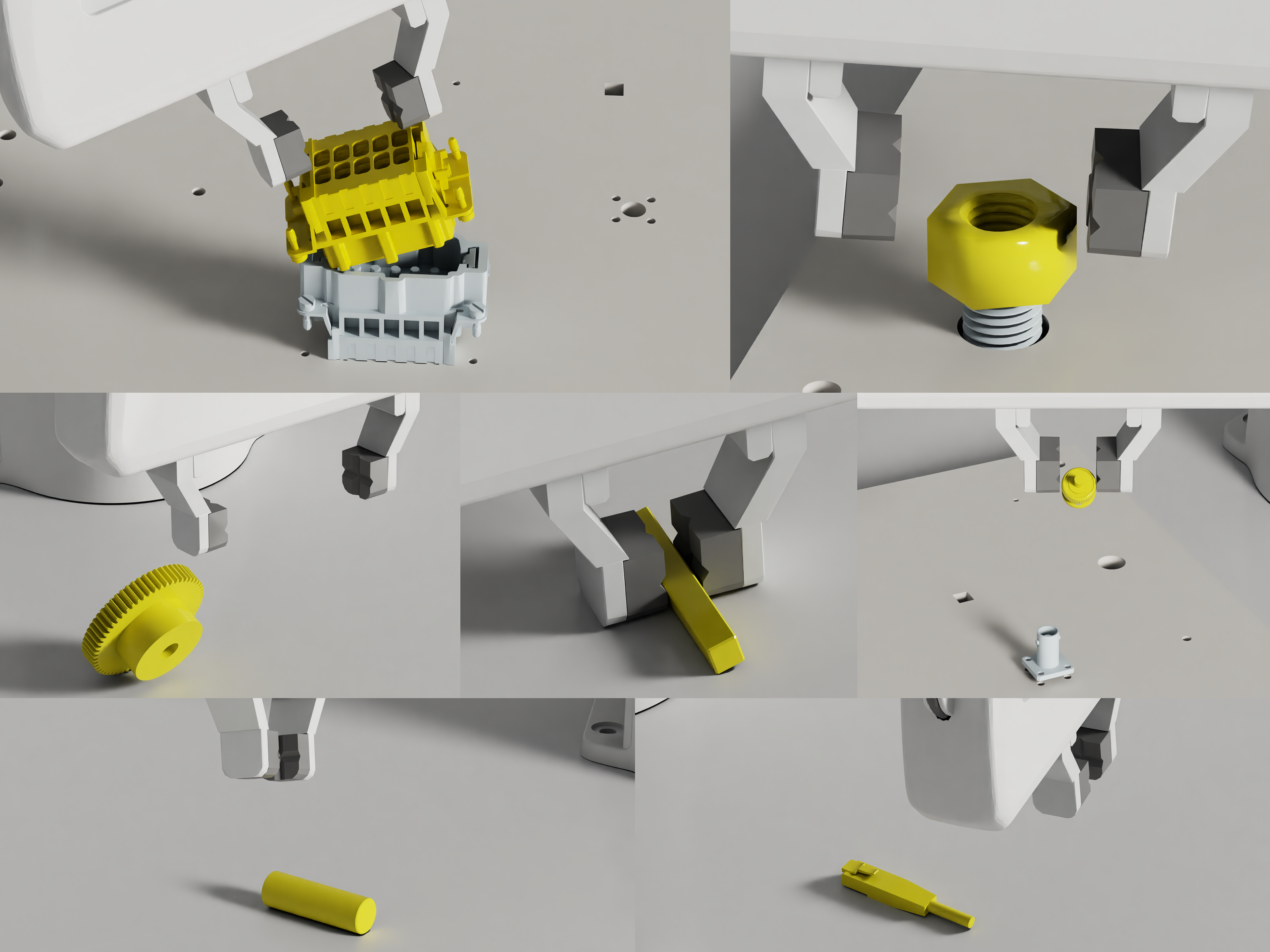}
    \caption{\textbf{Diverse assembly reset states.} Examples of resets across
    tasks, ranging from random object poses to in-hand grasps and partial
    assemblies.}
    \label{fig:reset-coverage}
\end{figure}

We adapt OmniReset~\cite{yin2026omnireset} to heterogeneous assembly by
caching the assembly identity together with each reset state:
\begin{equation}
    \mathcal{D}_{\mathrm{reset}}
    =\{(e_i,\bar{s}_i)\}_{i=1}^{M}.
\end{equation}
Here, $M$ is the total number of entries across all 20 assemblies.
The assembly identity $e_i$ selects the part and mating-geometry assets and their inertial properties, while $\bar{s}_i$ specifies the robot,
board, and movable-part state. Sampling an entry therefore selects
both the assembly and its initial configuration.
The cached bank spans ungrasped
workspace, pickup, transport, near-mating, and partially assembled
configurations (Fig.~\ref{fig:reset-coverage}), enabling GPU-parallel
initialization. These reset families
define regions of the training distribution, not a prescribed sequence; the
policy receives no stage identifier.

Let $\hat{p}_i$ be the empirical success rate over the most recent $L$ episodes
initialized from entry $(e_i,\bar{s}_i)$. Under beta sampling, entry $i$ is
selected with probability $q_i$:
\begin{equation}
    q_i\propto \varepsilon_q+
    \sqrt{\hat{p}_i(1-\hat{p}_i)},\qquad i=1,\ldots,M,
    \label{eq:beta-sampling}
\end{equation}
normalized over the entire bank. Sampling $i\sim\operatorname{Cat}(\mathbf q)$
selects an assembly--reset pair with $\rho(e_i,\bar{s}_i)=q_i$.
The kernel favors intermediate success rates, while $\varepsilon_q>0$
keeps every entry sampleable. Each episode's outcome updates $\hat{p}_i$
for its sampled entry.
The uniform ablation sets $q_i=1/M$; the bank, policy, reward, and
optimizer remain unchanged.
Algorithm~\ref{alg:training} summarizes training.

\subsection{Binary Full-Assembly Objective}

Across geometries, the assembly objective is the same: complete the assembly.
For each instance, an assembly-relative coordinate $\Phi_e(\alpha)$,
$\alpha\in[0,1]$, parameterizes the part in the mating frame, with
$\Phi_e(1)$ denoting its fully assembled pose. Let $d_{\mathrm{asm}}^e(s)$ be
the distance between the part's current relative pose and $\Phi_e(1)$.
For a single-pair episode, the binary task reward is
\begin{equation}
    r_t^{\mathrm{succ}}
    =
    \begin{cases}
        1, & d_{\mathrm{asm}}^e(s_{t+1})<\epsilon,\\
        0, & \text{otherwise}.
    \end{cases}
    \label{eq:sparse-reward}
\end{equation}
where $\epsilon=0.001\,\mathrm{m}$.
The success reward is issued once, and the episode terminates.
Intermediate progress receives no task reward.
We supplement the binary task reward with a control cost to discourage
large action magnitudes, rapid joint motion, and excessive torque.
This regularization biases learning toward smoother, lower-effort behaviors that are more compatible with safe deployment on physical robots, without providing information about assembly progress.
\begin{equation}
    r_t = \underbrace{r_t^{\mathrm{succ}}}_{\text{binary task reward}} - \underbrace{c\left(\lVert a_t\rVert_2^2 + \lVert \dot{q}_t\rVert_2^2 + \lVert \tau_t\rVert_2^2\right)}_{\text{control cost}}, \quad c \geq 0.
    \label{eq:full-reward}
\end{equation}
Here, $a_t$, $\dot{q}_t$, and $\tau_t$ are the action, robot joint velocity, and
applied joint torque. The control penalties are evaluated at every step,
with a single coefficient $c$ fixed across tasks. They do not measure
assembly progress; the binary task reward specifies successful completion.
Episodes end on assembly success, at horizon $T$, or when an arm joint
exceeds its hardware velocity or torque limit. %
In multi-pair episodes, the success reward remains zero until every active pair is
assembled, with no prescribed pair order or intermediate completion
reward.

\begin{figure*}[t]
    \centering
    \includegraphics[width=\textwidth]{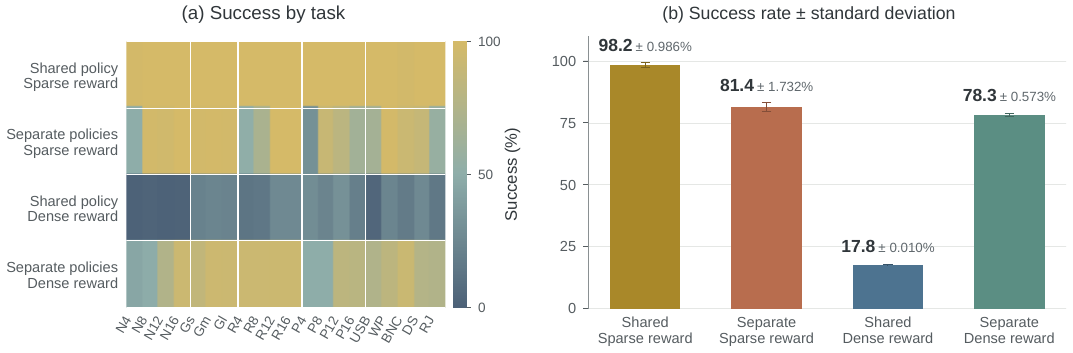}
    \caption{
    \textbf{Training across heterogeneous assembly pairs substantially improves learning from a sparse binary task reward.}
    At 64K environments across 25k iterations: (a) compares 20-task success rates for
    one shared policy versus 20 separately trained policies. The shared policy (first row)
    achieved above 98\% success rate across all tasks, including three precision assemblies that corresponding specialists rarely solve. Dense shaping does not exhibit the same benefit.
    Panel (b) shows overall success, note that the separate-policy success rate is average across all 20 per-task policies;
    \textit{Asset notation:} N, G, R, and P denote nuts, gears, rods, and pegs, respectively; numeric
    suffixes indicate nominal diameters in millimeters, while s, m,
    and l indicate gear size.
    "Sparse reward" denotes Eq. (6); "dense reward" additionally includes the shaping terms in Eq. (7).
    }
    \label{fig:heterogeneity-ablation}
\end{figure*}

\begin{figure}[!tb]
    \centering
    \includegraphics[width=\columnwidth]{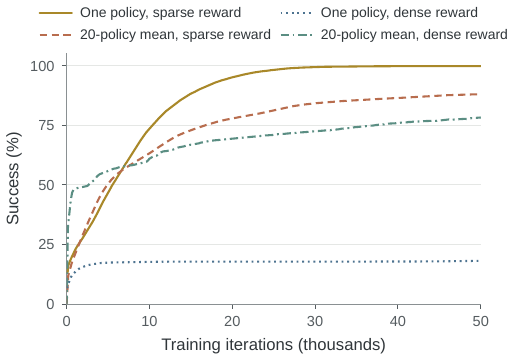}
    \caption{\textbf{Joint sparse training converges faster than separate
    policies.} Curves compare one shared policy trained on all 20 tasks with
    the mean of 20 policies trained separately, one per task.}
    \label{fig:heterogeneity-dynamics}
\end{figure}

\begin{figure}[!tb]
    \centering
    \includegraphics[width=\columnwidth]{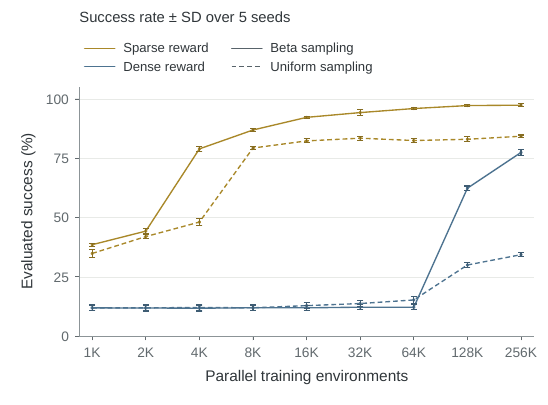}
    \caption{
    \textbf{Sparse task rewards benefit strongly from parallel experience.}
    Mean success $\pm$ standard deviation over 5 evaluation seeds for 
    25k-iteration checkpoints, with 32 trials per task per seed across 20
    assemblies. Color denotes reward; solid and
    dashed lines denote beta and uniform training-reset sampling.}
    \label{fig:parallel-scaling}
\end{figure}

\begin{figure*}[!tb]
    \centering
    \setlength{\tabcolsep}{2pt}
    \begin{tabular}{@{}cc@{}}
        \includegraphics[width=0.485\textwidth,trim=0 36 0 144,clip]{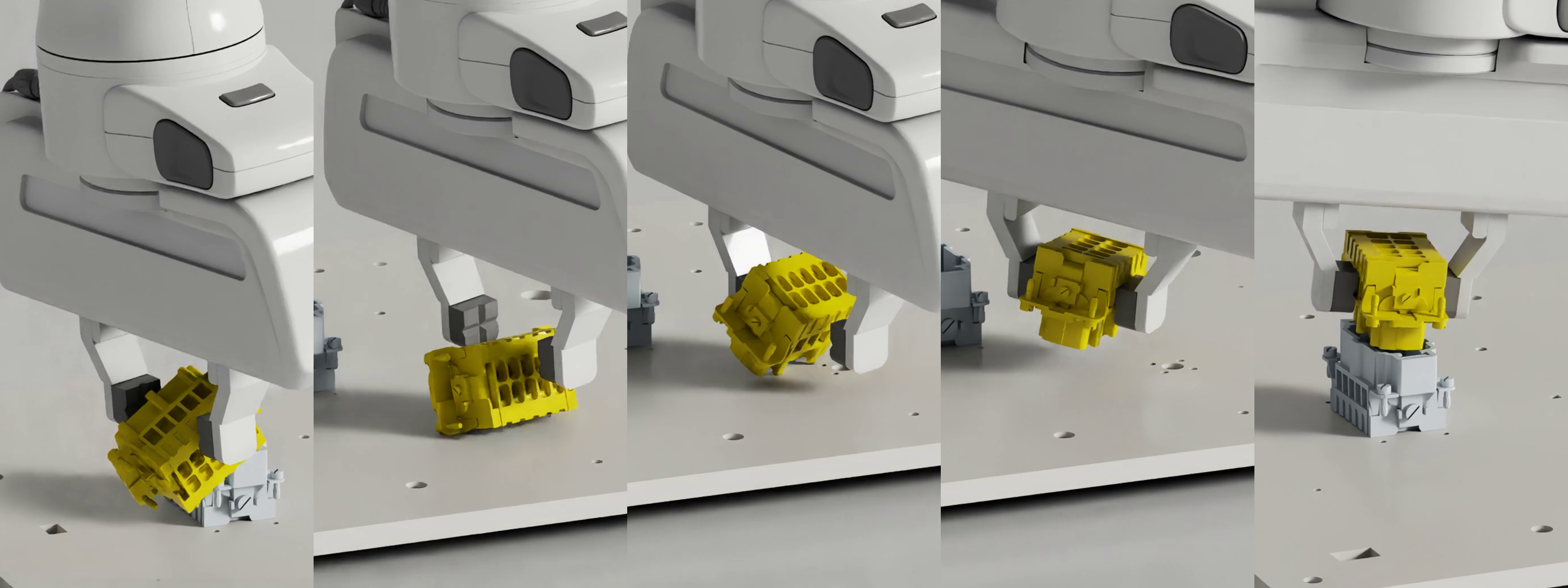} &
        \includegraphics[width=0.485\textwidth,trim=0 36 0 144,clip]{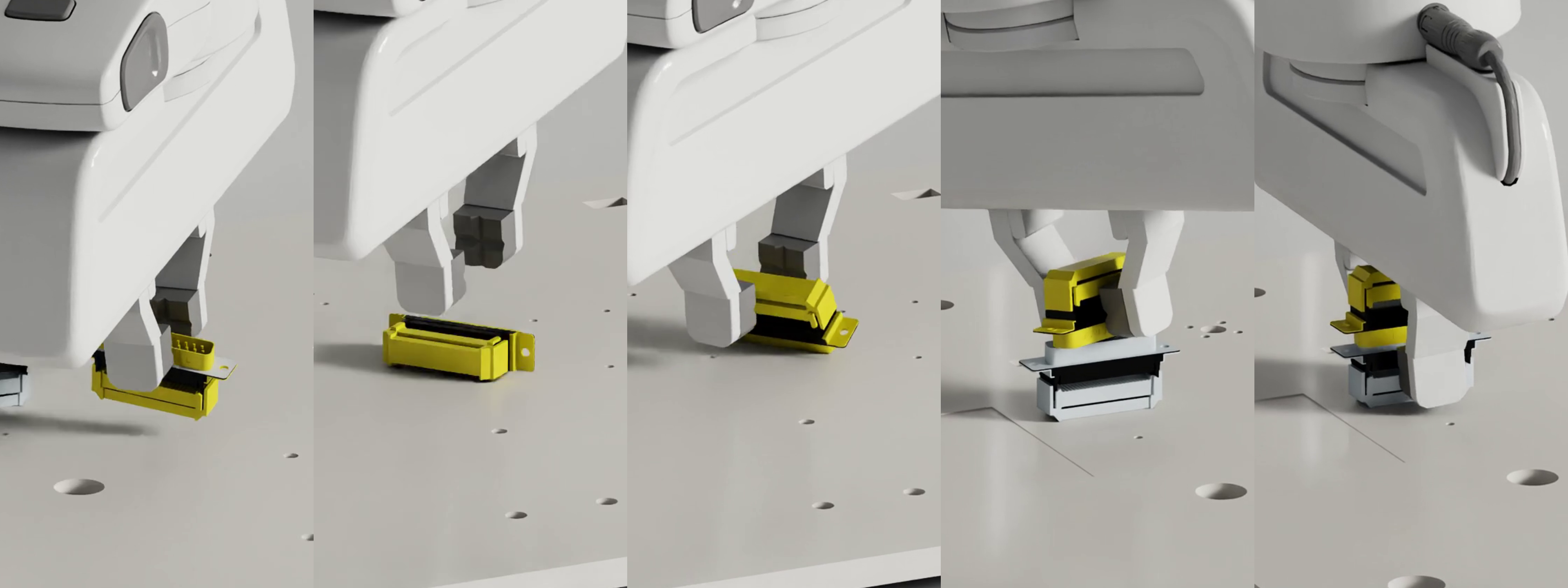} \\
        \footnotesize\sffamily\textbf{(a)} Board-assisted reorientation(waterproof) &
        \footnotesize\sffamily\textbf{(b)} Release-and-regrasp for alignment(d-sub) \\[2pt]
        \includegraphics[width=0.485\textwidth,trim=0 36 0 144,clip]{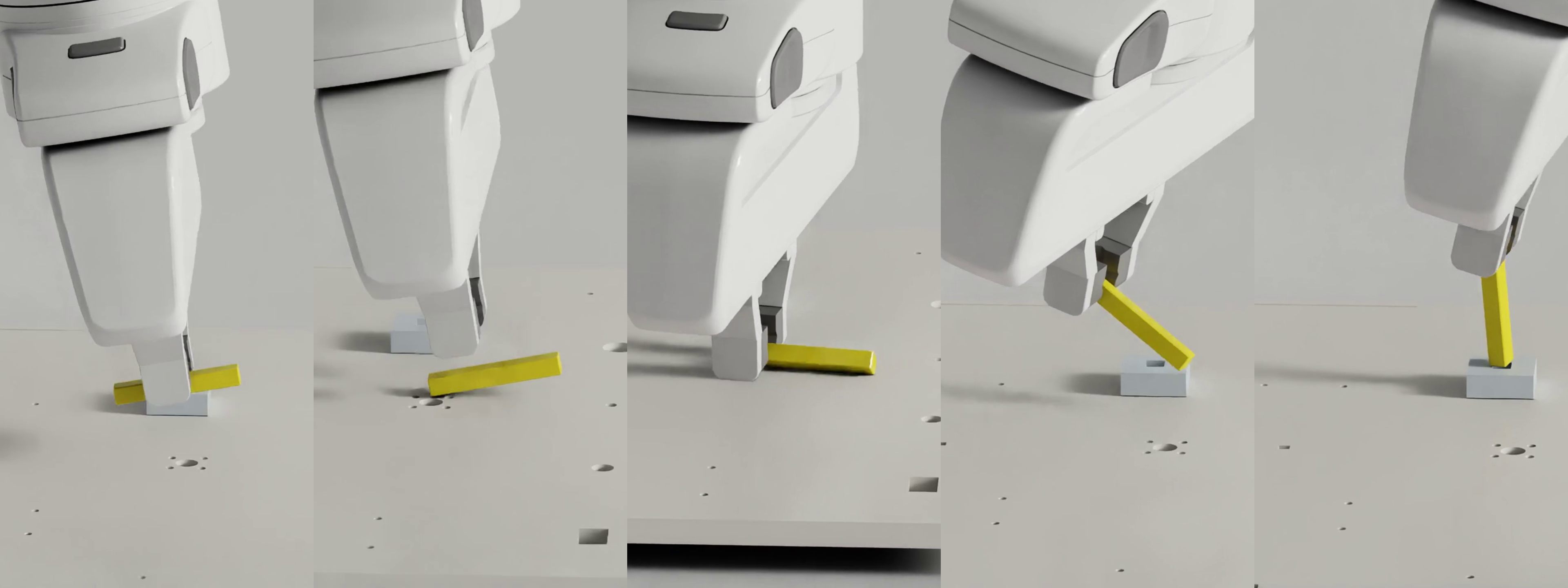} &
        \includegraphics[width=0.485\textwidth,trim=0 36 0 144,clip]{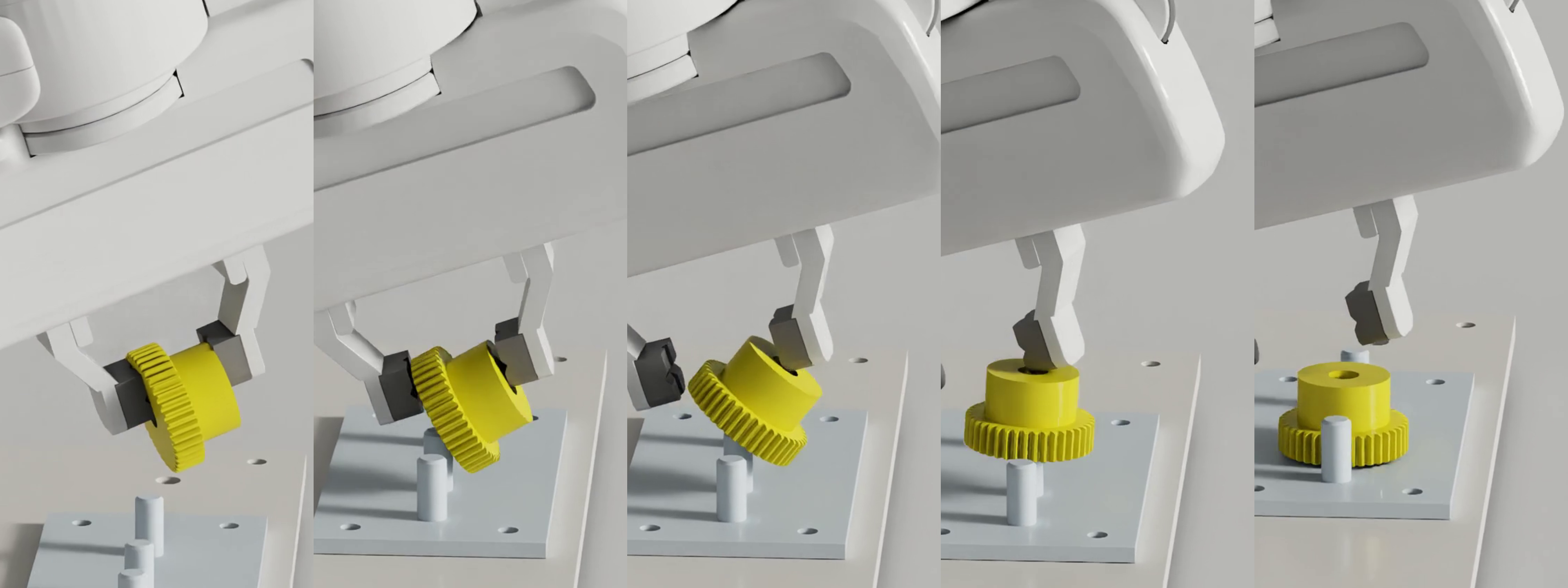} \\
        \footnotesize\sffamily\textbf{(c)} Center-to-tip grasp transfer(peg 8mm) &
        \footnotesize\sffamily\textbf{(d)} Side-caging for gear placement(medium gear) \\[2pt]
        \includegraphics[width=0.485\textwidth,trim=0 36 0 144,clip]{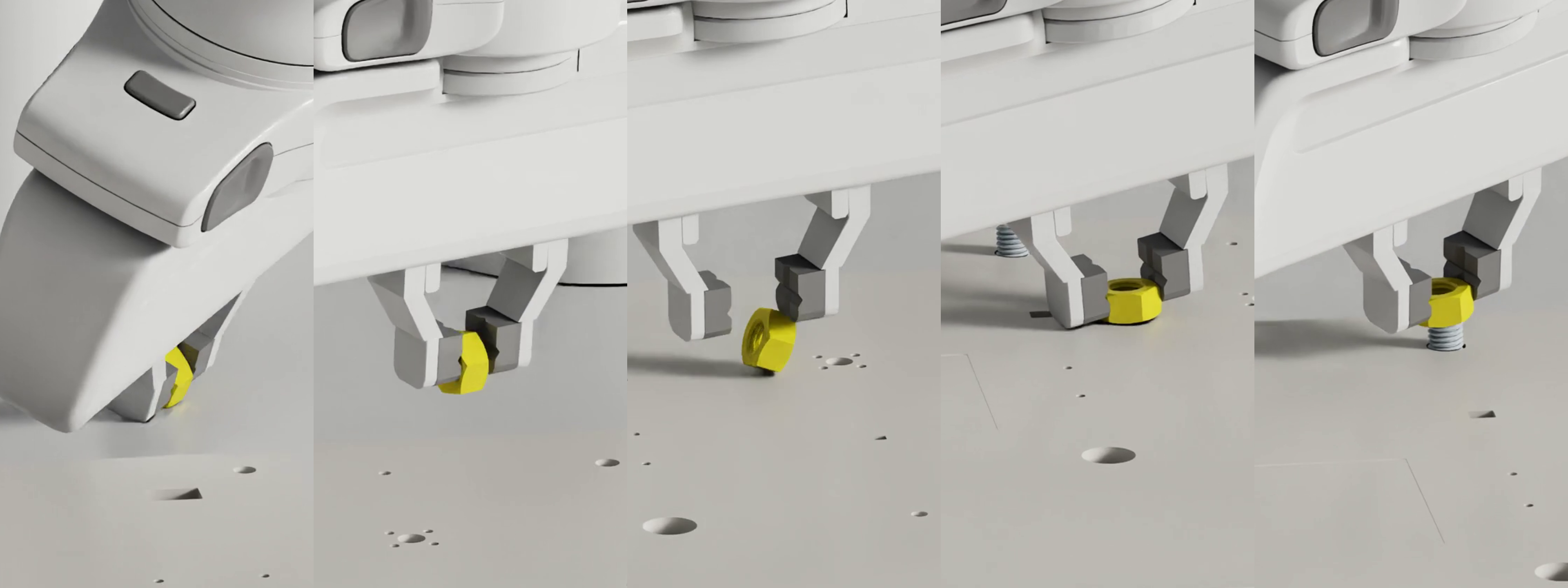} &
        \includegraphics[width=0.485\textwidth,trim=0 0 0 180,clip]{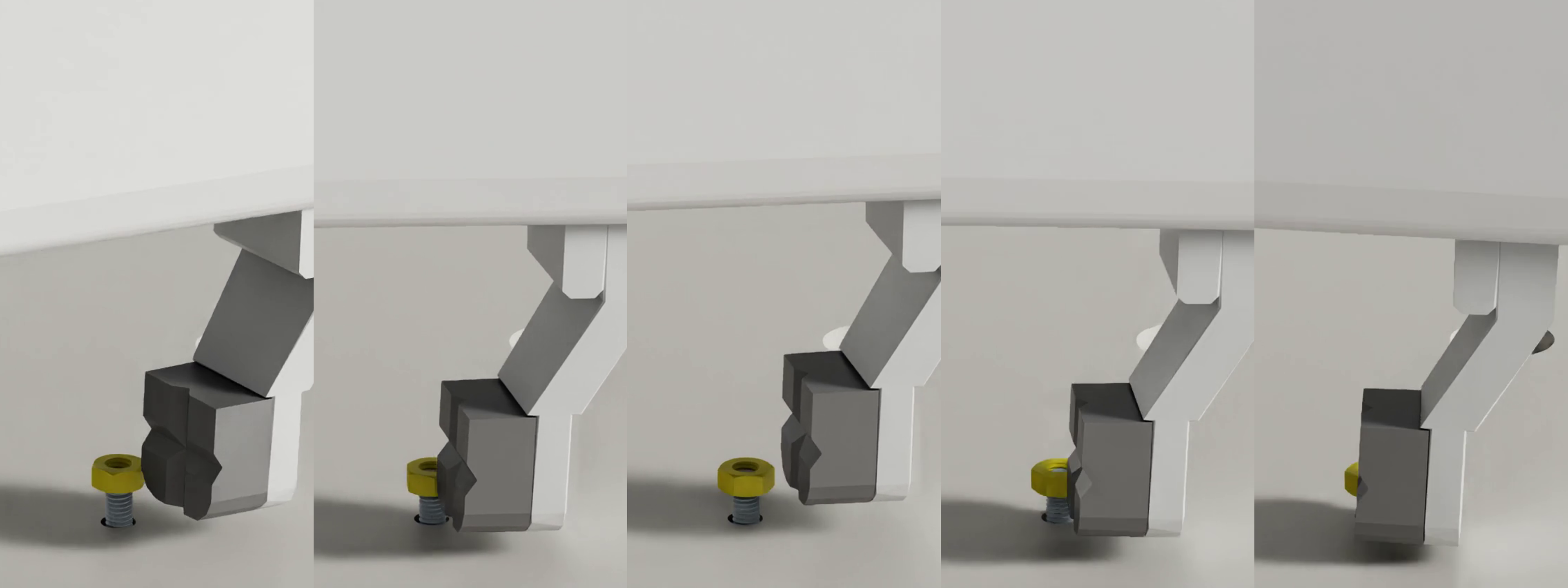} \\
        \footnotesize\sffamily\textbf{(e)} Edge-to-flat nut reorientation(nut m12) &
        \footnotesize\sffamily\textbf{(f)} One-sided tapping for threading(nut m4) \\[2pt]
        \includegraphics[width=0.485\textwidth,trim=0 36 0 144,clip]{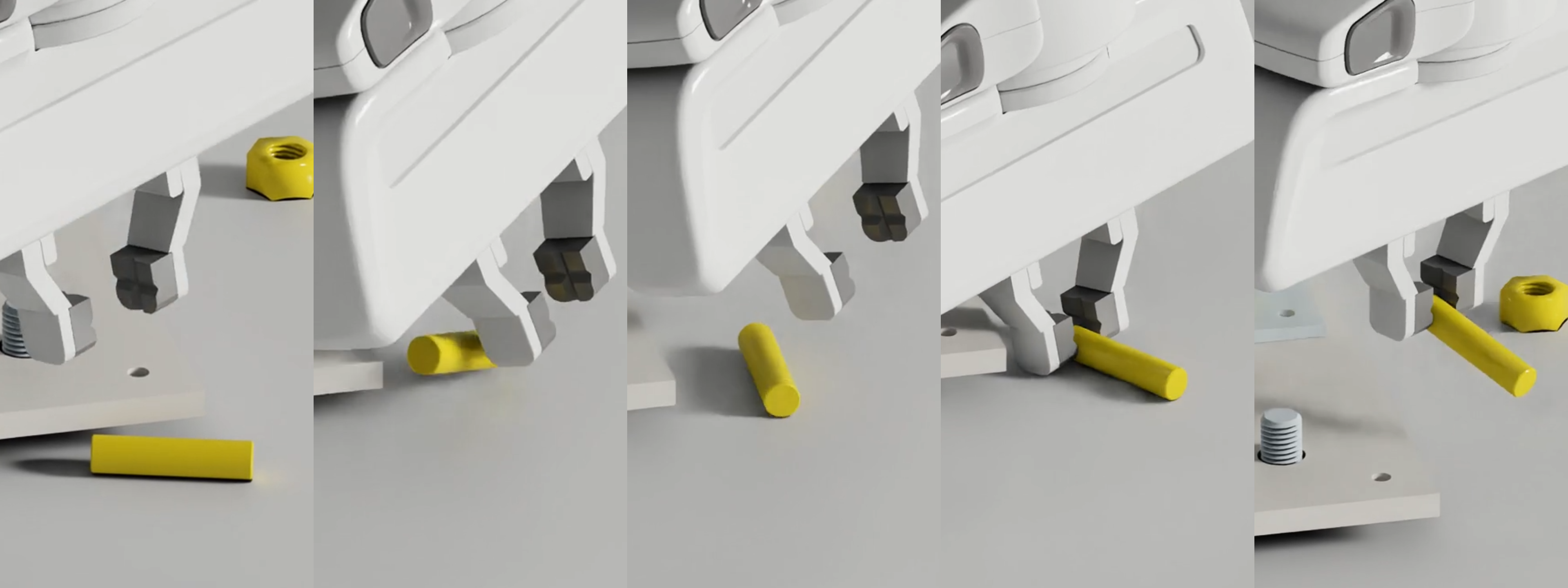} &
        \includegraphics[width=0.485\textwidth,trim=0 36 0 144,clip]{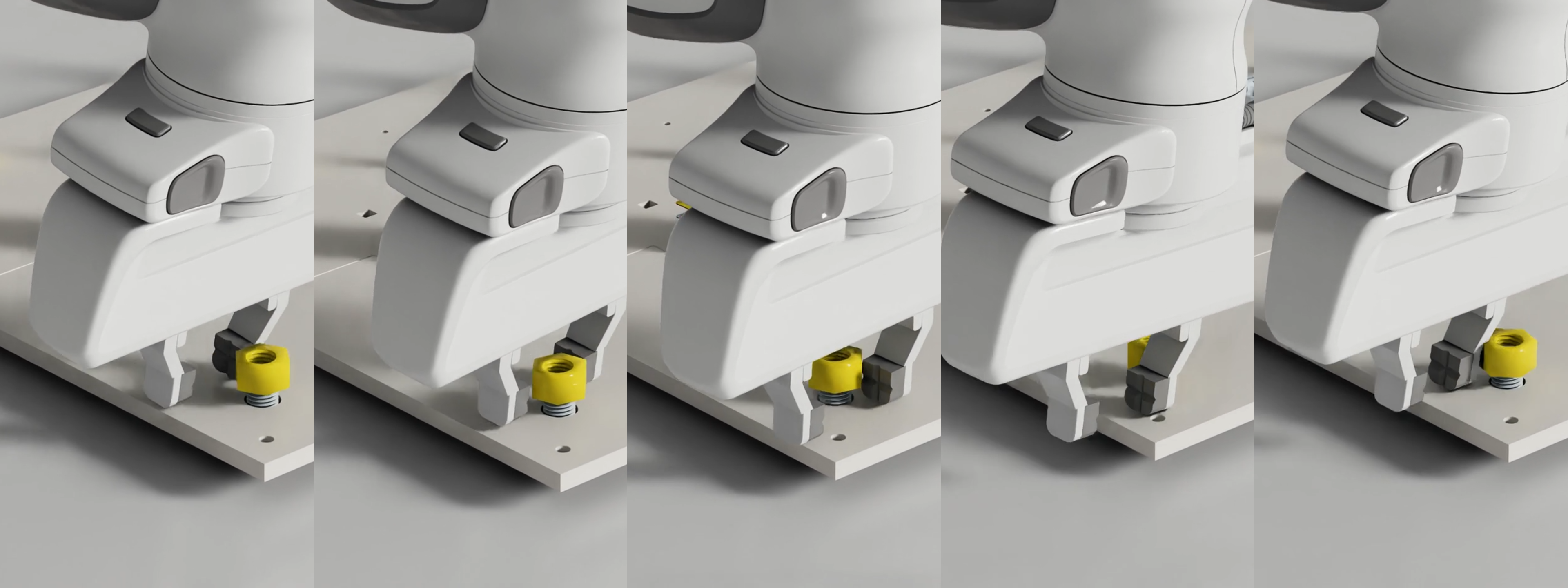} \\
        \footnotesize\sffamily\textbf{(g)} Shoveling to a better pickup position(rod 8mm) &
        \footnotesize\sffamily\textbf{(h)} Full-circle nut threading(nut m16)
    \end{tabular}
    \caption{\textbf{One shared policy discovers diverse contact strategies
    from a sparse binary task reward.} Each panel proceeds from left to right.
    (a) Waterproof connector: board contact helps reorient the part before insertion.
    (b) D-sub connector: the part is set down and regrasped for alignment with its socket.
    (c) 8 mm peg: the part is dropped from a center grasp and regrasped near one end,
    leaving its remaining length free for insertion.
    (d) Medium gear: the part pivots from a side grasp into a horizontal pose over its
    mounting post. (e) M12 nut: the nut is rolled from its edge onto a flat face before
    placement on the bolt. (f) M4 nut: repeated one-sided taps advance the nut along
    the thread. (g) 8 mm rod: the fingertip shovels the rod to a position where it is
    easier to lift. (h) M16 nut: the gripper moves around the nut, using fingertip
    contact to turn it along the bolt.}
    \label{fig:emergent-assembly-strategies}
\end{figure*}

\begin{figure*}[!tb]
    \centering
    \begin{tabular}{@{}c@{}}
        \includegraphics[width=\textwidth]{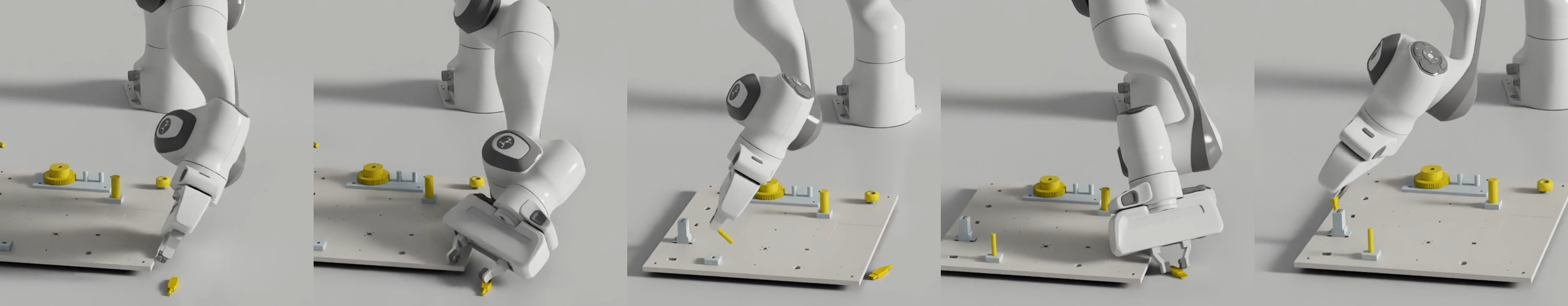} \\
        \footnotesize\sffamily\textbf{(a)} Bring USB-A closer, assemble the 8 mm rod, then return \\[4pt]
        \includegraphics[width=\textwidth]{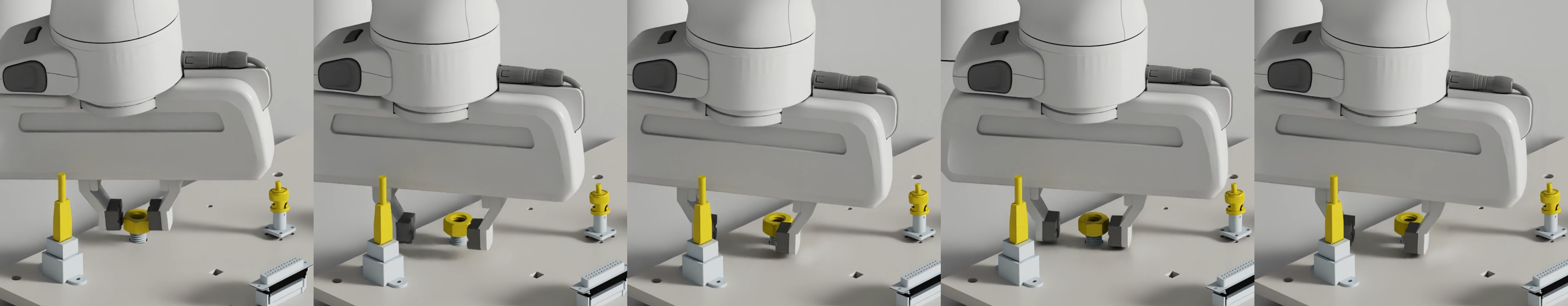} \\
        \footnotesize\sffamily\textbf{(b)} Making small turns threading M12 to avoid RJ45 collision
    \end{tabular}
    \caption{\textbf{Emergent coordination in multi-pair assembly.}
    (a) The policy brings the USB-A connector from a distant corner to a nearby
    working position, leaves it there while assembling the 8 mm rod, then
    returns to insert the connector. (b) During M12 nut threading, the
    end effector makes small turns to avoid touching the neighboring RJ45 connector.
    Each sequence proceeds from left to right.}
    \label{fig:long-horizon-multi-asset}
\end{figure*}

\subsection{Shared Policy and Robot Control}

We implement AlphaNIST in Isaac Lab~\cite{mittal2025isaaclab} with
GPU-parallel environments. Each PPO batch combines rollouts from
assembly--reset pairs sampled using Eq.~\eqref{eq:beta-sampling}.
The policy, reward, controller, and training configuration are shared
across all 20 tasks.
The policy observes the robot state and surface point clouds of the movable
part, mating geometry, and robot, without a discrete task identifier.
We use separate actor and critic SimBa networks
\cite{he2016deep,lee2025simba}, summarized in
Fig.~\ref{fig:policy-architecture}. Each combines a point-cloud embedding with
the proprioception and last action shown in the figure.

\begin{algorithm}[t]
    \caption{AlphaNIST Heterogeneous Assembly Learning}
    \label{alg:training}
    \small
    \begin{algorithmic}[1]
        \Require $\mathcal{D}_{\mathrm{reset}}=\{(e_i,\bar{s}_i)\}_{i=1}^{M}$;
        history length $L$; batch size $N$; shared $\pi_\theta,V_\phi$
        \State initialize each rolling success rate $\hat{p}_i\gets0$
        \For{each PPO iteration}
            \State $\mathcal{B}\gets\emptyset$
            \While{$|\mathcal{B}|<N$}
                \ForAll{completed episodes $n$}
                    \State $y_n\gets\mathbb{1}[\text{full assembly}]$
                    for entry $i_n$
                    \State $\hat{p}_{i_n}\gets$ mean of its latest $L$ outcomes
                \EndFor
                \State update reset probabilities
                $\mathbf q$ using Eq.~\eqref{eq:beta-sampling}
                \ForAll{environments $n$ requiring reset}
                    \State $i_n\sim\operatorname{Cat}(\mathbf q)$;
                    $(e_n,s_{n,0})\gets(e_{i_n},\bar{s}_{i_n})$
                \EndFor
                \State $a_{n,t}\sim\pi_\theta(\cdot\mid o_{n,t}),\quad\forall n$
                \State step all environments and append transitions to $\mathcal{B}$
            \EndWhile
            \State update $\pi_\theta$ and $V_\phi$ using standard PPO on $\mathcal{B}$
        \EndFor
    \end{algorithmic}
\end{algorithm}

We simulate a 7-DoF Franka Emika Panda with identified inertial and drivetrain
parameters and a parallel-jaw gripper with one driven and one mechanically
coupled passive finger. At $25\,\mathrm{Hz}$, the policy outputs a 7D relative joint position target ($q_t^{\mathrm{target}} = q_t + \Delta q_t$) and a 1D gripper command.
An implicit impedance controller uses the manufacturer's \textit{libfranka} stiffness
and damping profile with
hardware-matched armature and joint friction. Joint-speed limits are $2.175\,\mathrm{rad/s}$ for joints 1--4 and $2.61\,\mathrm{rad/s}$ for joints
5--7; torque limits are $87\,\mathrm{N\,m}$ and $12\,\mathrm{N\,m}$,
respectively. 
These constraints keep actions within a hardware-realistic
envelope, preventing the policy from exploiting unrealistic speeds or
torques. 
Similar hardware-aligned simulation has enabled transfer of
assembly policies to the same Franka Panda
platform~\cite{tang2023industreal,tang2024automate}.

\subsection{Distributed Training and Scale}

We scale synchronous PPO across 1--256K environments, with 4K per GPU
in multi-GPU runs. Each GPU samples its local reset bank using rolling
success estimates. Actor and critic gradients are averaged across GPUs
before each optimizer step. 
Each iteration collects 32 steps per environment and optimizes for 5
epochs with 4 minibatches each. Network architecture and PPO settings
are fixed across scales.

\section{Experiments}
\label{sec:experiments}
We organize the experiments around three questions:
\textbf{(Q1) Heterogeneity:} Under a binary task reward, how does joint
training compare with training each task separately?
\textbf{(Q2) Scaling:} How do parallel experience and reset-state sampling
affect learning from a binary task reward?
\textbf{(Q3) Reward formulation:} How does a binary task reward compare
with dense assembly-progress shaping as heterogeneous training scales?
We additionally explore multi-pair assembly and inspect the learned policy
for behaviors that arise without direct supervision.

For heterogeneity, we compare one policy trained across all 20 ATB1 tasks
with one separately trained policy per task, using matched per-run
interaction budgets and otherwise identical settings. We use 64K environments
per run to keep the comparison affordable; the shared policy reaches
98.2\% training success at this scale.

For parallel scaling, GPUs, environments, and rollout batch size increase
together while the task distribution, reward, policy, and environment remain
fixed. The sampling ablation compares against uniform sampling over the same reset
bank instead of beta sampling.
For the scaling evaluation, all checkpoints use stochastic actions and a common
frozen mixed-reset bank, sampled uniformly within each task. Success requires
full assembly within 15 seconds without a simultaneous safety failure.
The reward ablation adds two dense assembly-progress terms to
Eq.~\eqref{eq:full-reward}: 
\begin{equation}
    r_t^{\mathrm{dense}} = r_t +
    c_2 r_t^{\mathrm{reach}} + c_3 r_t^{\mathrm{align}},
    \qquad c_2,c_3 \geq 0,
    \label{eq:reward-ablation}
\end{equation}
where $r_t^{\mathrm{reach}}$ rewards end-effector proximity to the part 
centroid and $r_t^{\mathrm{align}}$ rewards alignment with the target assembly
orientation. These are common shaping terms in manipulation and assembly RL
\cite{narang2022factory,tang2024automate,yin2026omnireset}.
We set $c_2=c_3=0.01$, matching the coefficients used in
OmniReset~\cite{yin2026omnireset}.
Across all experiments, success rates cover all reset configurations.

\begin{table*}[!t]
\centering
\renewcommand{\arraystretch}{1.2}
\caption{\textbf{Success rate for multi-pair assembly} 
Each policy is trained and evaluated on the same number of active assembly pairs, from 1--10. 
AlphaNIST solves these tasks without a prescribed order or explicit task decomposition. 
For example, when there are 10 active pairs, the policy receives a success reward only when all 10 are in their assigned board slots. 
}
\label{tab:seq-success}
\begin{tabularx}{\textwidth}{l*{10}{>{\centering\arraybackslash}X}}
\toprule
\# Active\ Pairs & 1 & 2 & 3 & 4 & 5 & 6 & 7 & 8 & 9 & 10 \\
\midrule
Success Rate (\%) & 99 & 97 & 95 & 93 & 92 & 90 & 89 & 87 & 86 & 84 \\
\bottomrule
\end{tabularx}
\end{table*}

\subsection{Assembly-Pair Heterogeneity Enables Learning}
From randomly initialized states, separately-trained policies have $\leq$0.3\% success on M4 nut threading, as well as 4\,mm peg insertion and rod
insertion. The main obstacle is that the small parts are
difficult to grasp reliably from the table. 
In contrast, the shared
policy produces successful grasp-and-assembly rollouts on these same tasks
without task-specific changes. This result suggests that shared experience
can help with assemblies that are difficult to learn separately.
This benefit depends strongly on the reward formulation.
At their final checkpoints, the 20 separately trained dense-reward policies
reach a mean success rate of 78.3\%, ranging from 45.0\% on
M4 threading to 93.6\% on the medium gear. Joint dense training reaches 17.8\%
overall success at 25K iterations, with low completion across tasks
(Fig.~\ref{fig:heterogeneity-ablation}).
One possible explanation
is that dense shaping terms favor different partial solutions across geometries,
whereas the binary task reward gives a unified full-assembly objective for every task.

Joint training reaches higher success at lower total cost.
At 64K parallel environments across 25K iterations, the \textbf{shared policy trained with the sparse binary task reward reaches 98.2\% training success} in 410 GPU-hours
(Fig.~\ref{fig:heterogeneity-dynamics}), while the 20 separately trained policies average 81.4\% and consume 16.1K total GPU-hours,
39.3$\times$ more device-hours.

\subsection{Sparse Binary Task Rewards Benefit from Scale}
Performance with the sparse binary task reward improves rapidly with increased parallel experience (Fig.~\ref{fig:parallel-scaling}).
With beta sampling, evaluation success rises from
38.6\% at 1K environments to 78.9\% at 4K and 86.9\% at 8K.
With dense reward shaping, success remains at 11.8--12.0\% over 1K--4K environments.
Beta sampling provides an additional gain over uniform reset sampling. At 256K environments, success with the sparse binary task reward reaches \textbf{97.4\% with beta sampling} versus 84.3\% with uniform sampling; the corresponding results with dense reward shaping are 77.4\% and 34.3\%. While the advantage of the sparse task objective is already visible at modest scale, beta sampling provides further gains as
parallel experience increases.

\subsection{Emergent Strategies over Short Horizons}

The policy exhibits a broad range of dexterous strategies across
assemblies (Fig.~\ref{fig:emergent-assembly-strategies}), including grasp adaptation, extrinsic reorientation, and contact-driven assembly. It releases and reacquires parts when the initial grasp is unsuitable for insertion, uses surrounding geometry to change object orientation, and discovers intermittent tapping and pivoting motions.
These behaviors emerge within a single policy, without demonstrations or rewards specifying the individual
maneuvers.
For a peg, a successful pickup may not produce a grasp suitable for insertion; the
policy drops a center-held peg and regrasps it near one end, exposing enough
shaft length for complete insertion. For a nut, repeated taps on one side engage
and advance the thread, using intermittent contact to make assembly progress.

\subsection{Emergent Strategies over Long Horizons}
Beyond the main single-assembly experiments, we train policies with multiple
assembly pairs per episode. The binary task reward is issued only when every
active pair is assembled; episodes allow 15 seconds per pair and terminate upon success. 
Table \ref{tab:seq-success} shows the success rates for episodes with up to ten active pairs.
This multi-pair setting reveals coordination over substantially longer horizons (Fig.~\ref{fig:long-horizon-multi-asset}). The policy need not finish
assembling a pair before turning to another, and its approach to
one assembly changes based on neighboring parts. 
These planning-like behaviors---taking actions whose utility is realized later and adapting one assembly to the state of another---emerge without an explicit planner, policy switching, or prescribed sequence.

\FloatBarrier
\section{Limitations}
\label{sec:limitations}
First, all results are in simulation. Hardware-matched velocity and torque
limits constrain the learned motions, but do not establish robustness to
perception error, latency, or contact-model mismatch. 
Transfer and repeatability
will be evaluated on the physical board using sensor-derived observations and
randomized contact and latency models.
Second, while applying the 10-pair policy to successive groups of parts could offer a route to completing the full board, extending our multi-pair framework to 20 parts remains an exciting research goal. Achieving this at scale, under a single terminal task reward, will likely benefit from more efficient simulation and learning methods.

\section{Conclusion}
\label{sec:conclusion}
AlphaNIST shows that heterogeneous contact-rich assembly can be learned with a \emph{single} policy using a binary task objective, diverse reset coverage, and large-scale reinforcement learning.
In simulation, one policy learns all
20 ATB1 tasks without demonstrations, prescribed stages, or per-task tuning, while outperforming separately trained specialists on the most difficult small-part assemblies.
Across all tested scales, the binary task objective outperforms the evaluated dense reward shaping.
The learned policy discovers grasp adaptation, extrinsic reorientation, tapping, threading, and longer-horizon coordination without direct supervision.
In multi-pair assembly, it can prepare one part, complete another, and return to finish the first, without an intermediate assembly reward or a planner prescribing the sequence.

More broadly, these findings challenge the assumption that task complexity and diversity must be reduced to make robot learning tractable. Instead, they show that diverse experience under a shared objective can itself become a source of reusable dexterity and coordination. We view this work as a critical step toward general-purpose assembly policies that discover how to solve complex physical tasks, rather than requiring those strategies to be specified in advance.

\bibliographystyle{IEEEtran}
\bibliography{references}

\end{document}